\pdfoutput=1
\documentclass[letterpaper, 10 pt, journal, twoside]{IEEEtran}
\usepackage{cite}
\usepackage{url}
\usepackage{amsmath,amssymb,amsfonts}
\usepackage{algorithmic}
\usepackage{graphicx}
\usepackage{textcomp}
\usepackage{xcolor}
\usepackage{algorithm} 
\usepackage{float}
\usepackage{acronym}
\usepackage{booktabs}
\usepackage{siunitx}
\usepackage{threeparttable}
\usepackage[export]{adjustbox}
\DeclareSIUnit\px{px}

\acrodef{NeRFs}{Neural Radiance Fields}
\acrodef{SDFs}{Signed Distance Fields}
\acrodef{ESDF}{Euclidean Signed Distance Field}
\acrodef{ESDFs}{Euclidean Signed Distance Fields}
\acrodef{TSDF}{Truncated Signed Distance Field}
\acrodef{TSDFs}{Truncated Signed Distance Fields}
\acrodef{std}{standard deviation}
\acrodef{GANs}{Generative Adversarial Networks}
\acrodef{MAE}{mean absolute error}
\acrodef{MGD}{mean gradient difference}
\acrodef{PSNR}{peak signal-to-noise ratio}
\acrodef{SSIM}{structural similarity index measure}
\acrodef{NDT}{Normal Distribution Transform}
\acrodef{ICP}{Iterative Closest Point}

\makeatletter
\let\NAT@parse\undefined
\makeatother
\usepackage[hidelinks]{hyperref} 

\begin{document}

\title{Real-time Neural Dense Elevation Mapping for Urban Terrain with Uncertainty Estimations}
\author{Bowen Yang$^{1}$, Qingwen Zhang$^{1}$, Ruoyu Geng$^{1}$, \textit{Graduate Student Member, IEEE}, \\ Lujia Wang$^{2}$, \textit{Member, IEEE}, and Ming Liu$^{3}$, \textit{Senior Member, IEEE} %
\thanks{Manuscript received: August 4, 2022; Revised: October 31, 2022; Accepted: November 30, 2022. This paper was recommended for publication by Editor Markus Vincze upon evaluation of the Associate Editor and Reviewers’ comments. This work was supported by Guangdong Basic and Applied Basic Research Foundation, under project 2021B1515120032, awarded to Prof. Lujia Wang. (corresponding author: Lujia Wang.)} %
\thanks{$^{1}$Bowen Yang, Qingwen Zhang and Ruoyu Geng are with the Hong Kong University of Science and Technology, Hong Kong SAR, China. {\tt\small \{byangar,qzhangcb,rgengaa\}@connect.ust.hk}} %
\thanks{$^{2}$Lujia Wang is with the Hong Kong University of Science and Technology, Hong Kong SAR, China, and also with the Clear Water Bay Institute of Autonomous Driving. {\tt\small eewanglj@ust.hk}} %
\thanks{$^{3}$Ming Liu is with the Hong Kong University of Science and Technology (Guangzhou), Nansha, Guangzhou, 511400, Guangdong, China, and HKUST Shenzhen-Hong Kong Collaborative Innovation Research Institute, Futian, Shenzhen. {\tt\small eelium@ust.hk}} %
\thanks{Digital Object Identifier (DOI): see top of this page.}
}

\markboth{IEEE ROBOTICS AND AUTOMATION LETTERS. PREPRINT VERSION. ACCEPTED December~2022}%
{Yang \MakeLowercase{\textit{et al.}}: Neural Elevation Mapping for Urban Terrain with Uncertainty}


\maketitle

\begin{abstract}
Having good knowledge of terrain information is essential for improving the performance of various downstream tasks on complex terrains, especially for the locomotion and navigation of legged robots.
We present a novel framework for neural urban terrain reconstruction with uncertainty estimations.
It generates dense robot-centric elevation maps online from sparse LiDAR observations.
We design a novel pre-processing and point features representation approach that ensures high robustness and computational efficiency when integrating multiple point cloud frames.
A generative Bayesian model then recovers the detailed terrain structures while simultaneously providing the pixel-wise reconstruction uncertainty.
We evaluate the proposed pipeline through both simulation and real-world experiments. 
Our approach achieves high-quality terrain reconstruction with real-time performance on a mobile platform, and the uncertainty estimates may further benefit the downstream tasks of legged robots.
(See \href{https://kin-zhang.github.io/ndem/}{https://kin-zhang.github.io/ndem/} for more details.)
\end{abstract}

\begin{IEEEkeywords}
AI-Enabled Robotics, Legged Robots, Mapping.
\end{IEEEkeywords}

\section{Introduction}
\IEEEPARstart{R}{ecent} research presents promising progress in enabling mobile robots to operate in complex environments, where the scene perception quality is essential to task performance.
Although extensive research has been conducted on 3D scene reconstruction with detailed structures \cite{URF,neural3D,voxblox,vdbfusion}, the high computational costs limit their applications in various robotic applications like locomotion or local navigation.
Therefore, it's essential to obtain an efficient online scene representation, which is challenging due to the sparsity of observations, occlusions, and the inaccuracy of sensors and odometry \cite{hoeller2022neural}.

Due to the difficulty of directly recovering the scene from a single sparse and occluded observation frame, it's necessary to utilize the history frames.
Fankhauser \textit{et al.} \cite{elevation2014,elevation2018} gradually construct and update a robot-centric elevation map using the incoming observations.
The raw elevation map is further fed into a neural network to recover the occluded regions \cite{stolzle2022reconstructing}.
However, converting the raw observations into maps as the network inputs may lose information useful to the reconstruction task.
Hoeller \textit{et al.} \cite{hoeller2022neural} propose a neural scene reconstruction approach that adopts dense point clouds from depth cameras as the inputs and recurrently refines the previous results, while the recurrent structure may introduce the potential risk of accumulating reconstruction errors.

\begin{figure}[t]
\setlength{\abovecaptionskip}{0pt}
\centering
\includegraphics[width=3.10in]{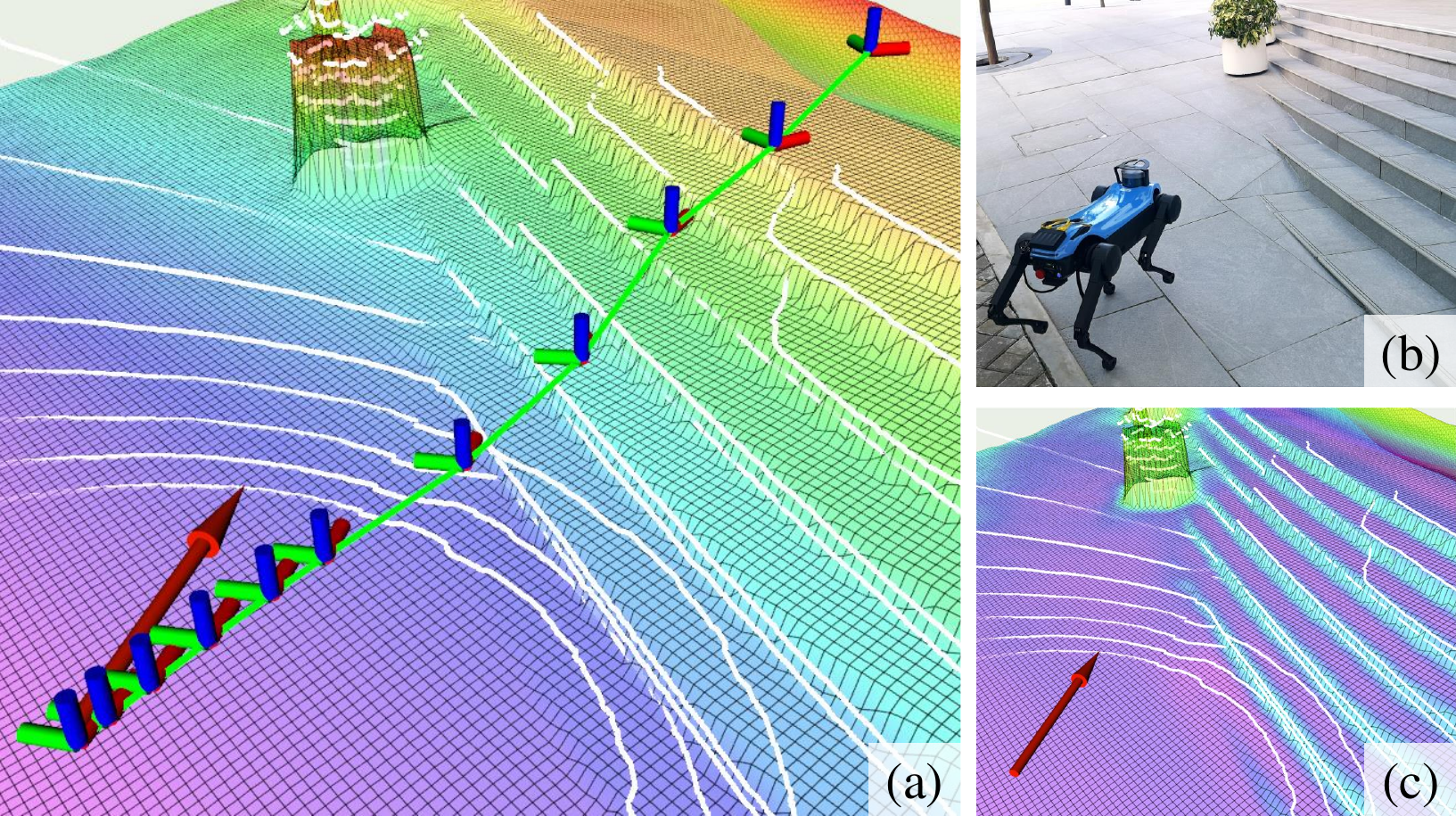}
\caption{We present an online neural dense elevation mapping approach that recovers detailed urban terrain structures from sparse and noisy $16$-channel LiDAR observations (a)(b), while additionally providing the terrain reconstruction uncertainty (c).
The framework can further benefit the downstream locomotion and navigation tasks, especially for legged robots.}
\label{fig:cover}
\vspace{-1.0em}
\end{figure}

To solve the challenges caused by occlusions and input inaccuracy, Miki \textit{et al.} \cite{elevationcupy} adopt traditional inpainting approaches and filter-based methods to generate and denoise dense elevation maps.
Due to the strong feature recovery capabilities, generative neural networks are introduced in \cite{hoeller2022neural,stolzle2022reconstructing,katyal2018occupancy,sharma2022occupancy} to eliminate the noises and generate rational features in the occluded regions.
However, directly using the inpainted maps in robotic applications may cause danger due to the lack of concrete terrain information in the occluded regions.
Therefore, providing the map reconstruction uncertainty is essential for the robustness of the downstream tasks.

This letter focuses on the urban terrain reconstruction task using elevation maps.
We present an end-to-end framework for online neural dense elevation mapping with measurable uncertainties using sparse LiDAR observations (Fig. \ref{fig:cover}).
A novel point cloud pre-processing approach maintains a group of statistical point features to represent the height distribution of the points inside each grid, enabling efficient integration of point frames.
A generative Bayesian model then returns dense elevation maps with detailed terrain structures from the noisy and sparse point features based on the prior knowledge of urban environments, outperforming the traditional approach \cite{elevationcupy} in several urban scenarios.
Compared with the state-of-the-art neural scene reconstruction approaches \cite{stolzle2022reconstructing,hoeller2022neural}, our approach additionally provides the reconstruction uncertainty for each grid, which may further improve the robustness of various downstream tasks.
Our main contributions include:
\begin{itemize}
    \item We propose an end-to-end framework for robust online neural terrain reconstruction to generate dense elevation maps from noisy and sparse LiDAR observations.
    \item We design a computational-efficient point cloud pre-processing approach to represent and maintain essential point features for high-quality mapping.
    \item We develop a generative Bayesian model that recovers detailed urban terrain structures and provides pixel-wise reconstruction uncertainty.
    \item We deploy our approach on an AGX Xavier for high-quality terrain mapping in real-world scenarios with real-time performance and present motivating examples to show its benefits to downstream robotic tasks.
\end{itemize}

\section{Related Work}    
    \subsection{3D Scene Reconstruction Approaches}
    Extensive research has been conducted on 3D scene reconstruction. 
    \ac{NeRFs} are constructed using multi-view images \cite{mildenhall2020nerf,Chen_2022_CVPR,Tancik_2022_CVPR} to represent large-scale urban environments \cite{URF,neural3D}.
    However, they normally require extensive observation of the scene for better performance.
    \ac{TSDFs} are widely used to represent detailed structures and can be used to incrementally build \ac{ESDFs} as proposed in Voxblox \cite{voxblox} for navigation.
    VDBFusion \cite{vdbfusion} improves the data structure of Voxblox \cite{voxblox} for better efficiency and more accurate ray-casting.
    Nevertheless, it still brings heavy computation and memory burdens to mobile devices and might be unsuitable for applications with online mapping requirements.
    
    Some approaches describe the volumetric occupancy or terrain height to simplify the calculation.
    Hornung \textit{et al.} \cite{octomap} propose OctoMap for probabilistic 3D occupancy estimation.
    Jia \textit{et al.} \cite{omu} develop an accelerator for real-time OctoMap building on embedded platforms.
    Stepanas \textit{et al.} \cite{ohm} adopt GPU for online occupancy map generation.

    Elevation maps are widely used as an efficient representation of terrain features.
    Fankhauser \textit{et al.} \cite{elevation2014,elevation2018} present a robot-centric probabilistic elevation mapping approach that also returns the uncertainty from the sensor noise and state estimation inaccuracy.  
    Miki \textit{et al.} \cite{elevationcupy} implement a GPU pipeline for efficient elevation mapping with various post-processing functions.
    However, these approaches may fail to recover dense and detailed terrain features when the observation is quite sparse.
    Our approach provides online dense elevation mapping from sparse LiDAR frames using a neural network.
    Different from the uncertainty in \cite{elevation2014,elevation2018}, our approach additionally considers the uncertainty from map inpainting as it generates dense maps, which may better benefit the downstream tasks for robust performance.
    
    \subsection{Scene Completion and Map Inpainting}
    Various map inpainting and scene completion methods leverage data-driven approaches or the prior information of terrain structures to solve the occlusion problem and optimize the reconstruction results.
    Existing works use learning-based approaches for 2D occupancy maps inpainting \cite{wei2021occupancy,sharma2022occupancy}, or 3D semantic scene completion \cite{song2017semantic,zhang2018semantic}.
    Hoeller \textit{et al.} \cite{hoeller2022neural} implement an online neural scene reconstruction method to generate 3D occupancy voxels for structured terrains using point frames from depth cameras.
    Different from \cite{hoeller2022neural}, our approach generates 2.5D elevation maps using sparse LiDAR frames and also provides the reconstruction uncertainty.
    
    Some recent work focuses on online elevation map inpainting.
    Miki \textit{et al.} \cite{elevationcupy} provide various traditional inpainting methods such as Telea \cite{telea2004image}, Navier-Stokes \cite{ebrahimi2013navier}, and minimum filter \cite{jenelten2022tamols}, and perform plane segmentation to recover the empty grids and extract geometric surfaces.
    Stölzle \textit{et al.} \cite{stolzle2022reconstructing} adopt self-supervised learning to train a neural network that takes raw elevation maps as input and recovers the occluded regions.
    We present an end-to-end framework that efficiently pre-processes the raw sparse LiDAR observations to obtain the point features and then adopts a generative Bayesian model for real-time local dense elevation mapping. 
    Our approach better recovers the details of urban terrain structures compared with the traditional inpainting methods \cite{telea2004image,jenelten2022tamols} in \cite{elevationcupy} and further provides the reconstruction uncertainty compared with \cite{stolzle2022reconstructing}.

\begin{figure}[b]
\setlength{\abovecaptionskip}{0pt}
\centering
\vspace{-1.0em}
\includegraphics[width=3.00in]{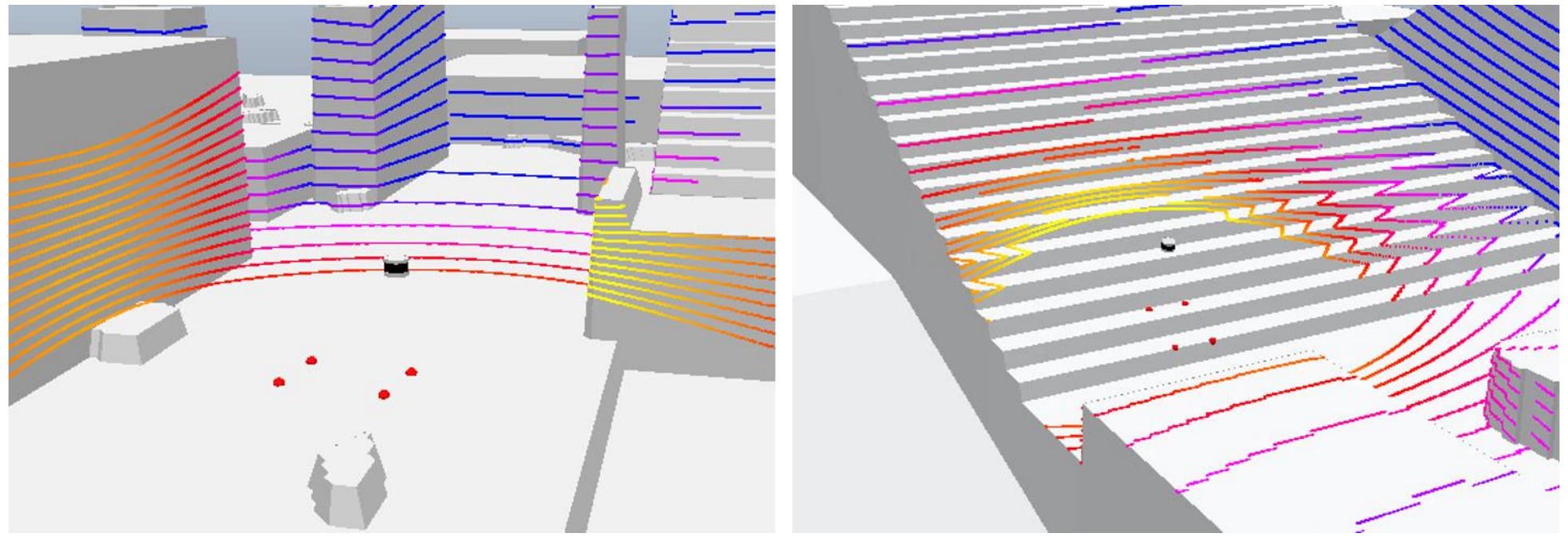}
\caption{Examples of the simulation environment. The generated maps contain diverse types of urban terrains. A $16$-channel LiDAR moves and observes the environment. The LiDAR's orientations change with the local topography considering the robot's foot configuration (red balls).}
\label{fig:sim_env}
\end{figure}

\begin{figure*}[ht]
\setlength{\abovecaptionskip}{0pt} 
\centering
\includegraphics[width=6.45in]{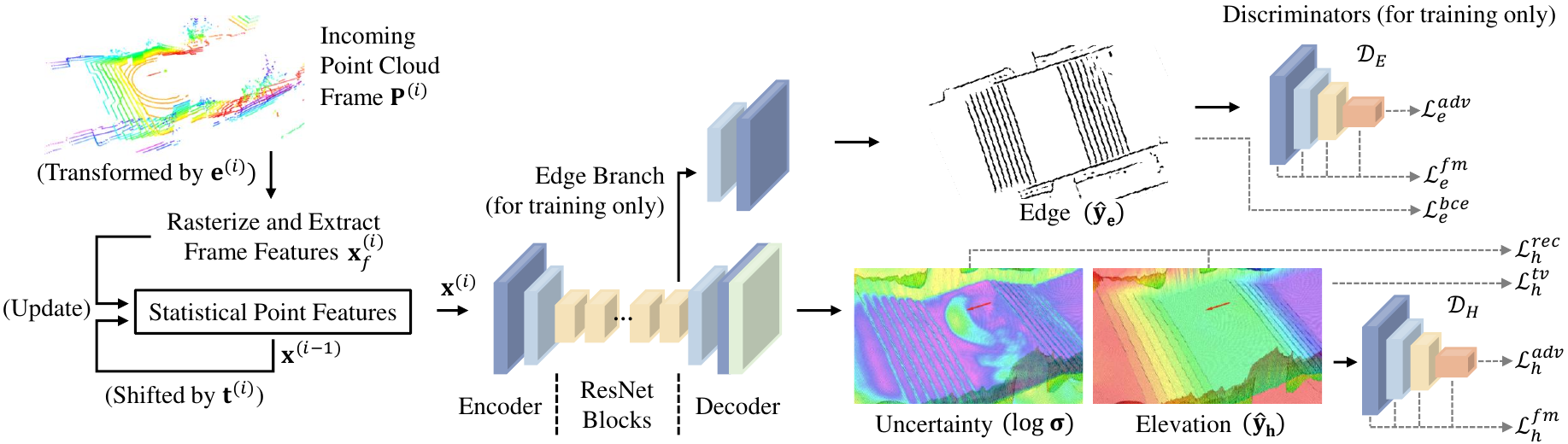}
\caption{The proposed elevation mapping pipeline. It first pre-processes the observed sparse point clouds to update the statistical point features. A generative Bayesian model then encodes the current point features to generate the dense elevation map and the pixel-wise reconstruction uncertainty while simultaneously returning a binary edge map for multitask learning in the training phase. Two discriminators are adopted to guide the learning of elevation map and edge map generation using adversarial loss $\mathcal{L}^{adv}$ and feature matching loss $\mathcal{L}^{fm}$ (See Section \ref{sec:training} for details of the loss terms).}
\label{fig:arch}
\vspace{-1.0em}
\end{figure*}

\section{Methodology}
This section explains our problem formulation and how we design the dataset generation environment, data pre-processing approach, network architecture, and training objectives.
    \subsection{Problem Formulation}
    We focus on the online mapping scenarios, where the mobile robot continuously observes the static environment and receives sparse 3D LiDAR point clouds $\mathbf{P}=\{\mathbf{p}_k|\mathbf{p}_k=(x_k,y_k,z_k)^T \in \mathbb{R}^3\}$.
    The terrains are represented as robot-centric 2D grid maps $\mathbf{y}_h$, where each grid contains a height value of that location.
    Our elevation mapping process is to obtain dense height estimations $\mathbf{\hat{y}}_h$ that recover the terrain features given sequential LiDAR frames $\{\mathbf{P}^{(i)}\}$ and robot's motion estimates $\mathbf{e}^{(i)}=(\mathbf{R}^{(i)}, \mathbf{t}^{(i)}), \ \mathbf{R} \in \mathbb{R}^{3 \times 3}, \ \mathbf{t} \in \mathbb{R}^{3}$.
    
    Generating dense elevation maps to represent detailed terrain features is an ill-posed problem due to observation inaccuracy and occlusions.
    Therefore, we specifically concern the urban scenarios with the assumption that the terrains are structured with sharp edges, flat surfaces, or smooth slopes.
    The point frame $\mathbf{P}^{(i)}$ at time $i$ is processed to maintain $\mathbf{x}^{(i)}$ which is a 7-channel feature map containing the statistical features of the history frames (explained in Section \ref{sec:features}).
    Then a generative model recovers the urban terrain features and generates dense maps.  
    The concept of Bayesian learning \cite{kendall2019geometry} is introduced to measure the mapping uncertainty.

    \subsection{Dataset Generation} \label{sec:simu_env}
    We prepare $\num{100}$ urban maps in simulation (Fig. \ref{fig:sim_env}).
    Each map is $\qtyproduct[product-units=single]{24 x 24}{\metre}$ with $\SI{60}{\percent}$ being flat grounds, $\SI{10}{\percent}$ being stairs, $\SI{10}{\percent}$ being slopes, and $\SI{20}{\percent}$ being boxes or irregular steps to represent other obstacles.
    The terrain generation parameters are uniformly sampled in pre-specified ranges: $[\num{0.20}, \num{0.35}]\unit{\metre}$ and $[\num{0.08}, \num{0.25}]\unit{\metre}$ for stairs' width and height, $[\num{0.18}, \num{0.60}]\unit{\radian}$ for the slope angle, $[\num{0.08}, \num{3.20}]\unit{\metre}$ and $[\num{-0.50}, \num{2.00}]\unit{\metre}$ for the side length and height of boxes and steps.
    A $16$-channel LiDAR is placed $\SI{0.50}{\metre}$ above the ground using the configuration of a Jueying Mini \cite{Jueying} quadrupedal robot (Fig. \ref{fig:cover}(b)).
    We move the robot along a predefined trajectory around the map center with a linear velocity uniformly sampled in $[\num{0.0}, \num{1.0}]\unit[per-mode = symbol]{\meter\per\second}$. 
    We calculate the robot's foot positions using the robot configuration and the local topography, and obtain the corresponding LiDAR positions and orientations, which are slightly perturbed to simulate the body vibrations from locomotion.
    The ground truth height maps $\mathbf{y}_h$ are $\qtyproduct[product-units=single]{5 x 5}{\metre}$ robot-centric patches with a resolution of $\SI{0.04}{\metre}$.
    We also generate the ground truth binary edge maps $\mathbf{y}_e$ using the Canny edge detector as done in \cite{kendall2019geometry} for multitask learning \cite{caruana1997multitask} in Section \ref{sec:arch}.
    
    The observations are augmented to reduce the sim-to-real gap \cite{hoeller2022neural}.
    When a point cloud is received, we add uniformly sampled values in $[\num{-0.02}, \num{0.02}]\unit{\metre}$ on point coordinates to mimic the sensory noise.
    The ground truth robot odometry is also perturbed in each step with a uniformly sampled translation in $[\num{-0.02}, \num{0.02}]\unit{\metre}$ and a rotation in $[\num{-0.04}, \num{0.04}]\unit{\radian}$ to simulate the inaccuracy from pose estimation.
    
    We define a grid cell inside the mapping region to be observed if at least one data point is found at that cell during the robot motion.
    Since the observations right after initialization are too sparse for meaningful results, we calculate the ``observation ratio" of a patch, which is the number of observed grids divided by the total number of grids in that patch, to filter out invalid samples.
    A sample will be dropped if the patch's observation ratio is less than $\SI{25}{\percent}$.
    
    \subsection{Representation of Point Features} \label{sec:features}
    
    We design an efficient data pre-processing approach to represent the distributions of the point height values in each grid, as they reflect the terrain structures. 
    For example, the height values of points on the flat ground gather around the height of the plane, which approximately follows a Gaussian distribution.
    While on a vertical edge, the points may distribute uniformly along the vertical surface.
    We adopt Gaussian models to represent the point distribution properties and obtain the statistical features for the points in each grid to generate a 7-channel feature map $\mathbf{x}^{(i)}$ at time $i$ as the network input. 
    The statistical features (in upper case) contain the number of points $C^{(i)}$, the mean and variance of height values $E^{(i)}(Z)$ and $Var^{(i)}(Z)$, and the mean and variance of the maximum and the minimum height over time $E^{(i)}(Z_{max})$, $Var^{(i)}(Z_{max})$ and $E^{(i)}(Z_{min})$, $Var^{(i)}(Z_{min})$. 
    
    As shown in the left part of Fig. \ref{fig:arch}, when we receive a point frame $\mathbf{P}^{(i)}$, we first transform it to the map coordinate with the motion estimates $\mathbf{e}^{(i)}$ and rasterize it inside the mapping region to extract grid-wised features (in lower case) for this single frame. Each grid in the frame feature $\mathbf{x}_f^{(i)}$ contains the number of points $c^{(i)}$, the sum of height values $\sum z^{(i)}$, the sum of the squared height values $\sum (z^2)^{(i)}$, and the maximum and minimum height values $z_{max}^{(i)}$ and $z_{min}^{(i)}$.
    The empty grids are filled with zeros. 
    Next, the previous statistical features $\mathbf{x}^{(i-1)}$ are shifted by the translational terms of $\mathbf{e}^{(i)}$ to compensate for the ego-motion and then updated using $\mathbf{x}_f^{(i)}$. 
    For instance, the mean of the height values is updated as the overall sum of height values divided by the new total number of points:
    \begin{gather}
        C^{(i)} = C^{(i-1)} + c^{(i)}, \\
        E^{(i)}(Z) = \frac{C^{(i-1)} \times E^{(i-1)}(Z) + \sum z^{(i)}}{C^{(i)}},
    \end{gather}
    and based on the relationship $Var(z)=E(z^2)-E^2(z)$, the variance is updated as:
    \begin{gather}
        E^{(i-1)}(Z^2) = Var^{(i-1)}(Z) + (E^2)^{(i-1)}(Z), \\
        \sum(Z^2)^{(i)} = C^{(i-1)} \times E^{(i-1)}(Z^2) + \sum(z^2)^{(i)}, \\
        Var^{(i)}(Z) = \frac{\sum(Z^2)^{(i)}}{C^{(i)}} - (E^2)^{(i)}(Z).
    \end{gather}
    The mean and the variance values of $Z_{max}$ and $Z_{min}$ are obtained in the same way.
    The LiDAR frames can be pre-processed through parallel computing and the point features are updated with simple rules. 
    That's how we achieve high efficiency on GPU-enabled mobile devices.
    
    Although combining multiple LiDAR frames suffers from sensory noise and odometry inaccuracy, the maintained point distributions will converge to stable states as the number of observations increases, assuming that the odometry drifts are negligible in local mapping scenarios.
    This also improves the stability of neural map generation. 
    One side effect is that the influence of new frames will gradually decrease as the number of points accumulates.
    Therefore, we append a final step of the update that decays the point count $C^{(i-1)}$ of a grid with a factor $\gamma$ and limits the number of points with $C_{max}$ if there are new points observed at that grid:
    \begin{gather}
        C^{(i)} = \min(C_{max}, \ \gamma C^{(i-1)} + c^{(i)}),
    \end{gather}
    where we set the maximum point count $C_{max}=100$ and $\gamma=0.90$ to encourage the updates using new data frames.

    \subsection{Network Architecture} \label{sec:arch}
    One of our major objectives is to recover detailed terrain structures with high-quality edges because edges contain important terrain features that are quite essential to improve the safety and efficiency of downstream tasks. 
    As explained in \cite{nazeri2019edgeconnect}, traditional image inpainting pipelines suffer from blur or artifacts due to the difficulties in recovering the high-frequency component.
    In this case, they present EdgeConnect which sequentially adopts an edge generator and an image completion network to separately conquer the edge recovery and image inpainting.
    We reference the architecture of EdgeConnect while adopting multitask learning \cite{caruana1997multitask} for simultaneous edge generation and elevation mapping.
    This assists the model in maintaining high-frequency components in the bottleneck features while achieving higher inference speed compared with the serial structure of EdgeConnect.
    
    Fig. \ref{fig:arch} presents our pipeline and network architecture.
    The maintained point features $\mathbf{x}^{(i)}$ are first passed into convolutional layers for encoding, which are then down-sampled before entering $6$ ResNet blocks for feature extraction. 
    Each ResNet block contains a dilated convolutional layer for larger receptive fields and a normal convolutional layer for output.
    Two decoding blocks respectively up-sample the bottleneck features and generate the binary edges $\mathbf{\hat{y}}_e$ and height map $\mathbf{\hat{y}}_h$.
    Based on the heteroscedastic aleatoric uncertainty model in \cite{kendall2019geometry}, our height map generation block returns an additional channel that contains a log scale variance map $\log \boldsymbol{\sigma}$ which is integrated into the reconstruction loss in Section \ref{sec:training} and jointly optimized with the height map.
    The model is encouraged to return high $\log \boldsymbol{\sigma}$ values in the places where it's difficult to reconstruct the terrain due to strong noises or occlusions, and thus it reflects the reconstruction uncertainty.
    Two discriminators with convolutional layers $\mathcal{D}_E$ and $\mathcal{D}_H$ generate 4-time down-sampled binary classification maps to guide the training of edge and height map generation.
    During deployment, only the height map branch in the generator is activated for fast inference. 
    Because the network is fully convolutional, it can deal with different input sizes to adapt to the mapping requirements of various downstream tasks.
    
    \begin{figure*}[t]
    \setlength{\abovecaptionskip}{0pt}
    \centering
    \includegraphics[width=6.80in]{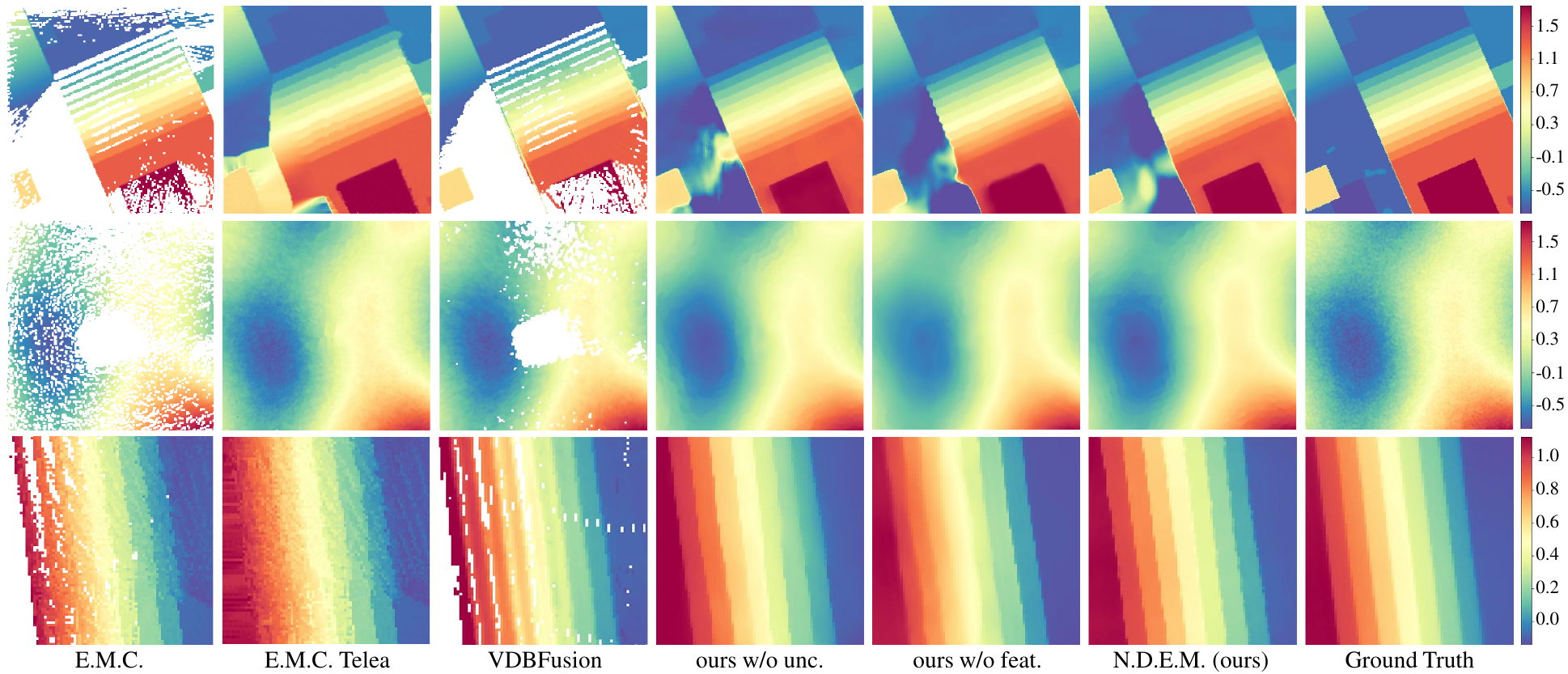}
    \caption{Visualized examples of the elevation mapping experiments in a simulated \textit{urban} scenario (the first row, $\qtyproduct[product-units=single]{8 x 8}{\metre}$), simulated \textit{hill} scenario (the second row, $\qtyproduct[product-units=single]{8 x 8}{\metre}$), and on real-world stairs (the third row, $\qtyproduct[product-units=single]{3.2 x 3.2}{\metre}$) using different approaches, where different colors indicate the height values. For \textit{urban} terrains and real-world stairs, our approach (N.D.E.M.) provides accurate dense elevation maps with high reconstruction quality and can further recover detailed terrain structures by introducing uncertainty estimation and adopting our statistical point features. In the \textit{hills} scenario, our approach can still recover the terrain in a global view. However, all of our models fail to recover the local features and wrongly generate smooth and step-like terrains.}
    \label{fig:baseline_vis}
    \vspace{-1.0em}
    \end{figure*}
    
    \subsection{Training Objectives} \label{sec:training}
    During the training process, the edge generation is optimized using a pixel-wise binary cross-entropy loss $\mathcal{L}_e^{bce}$:
    \begin{equation}
        \mathcal{L}_e^{bce}=-\sum \left(\mathbf{y}_e\log \mathbf{\hat{y}}_e + (1-\mathbf{y}_e)\log(1-\mathbf{\hat{y}}_e) \right).
    \end{equation}
    For height map generation, we use the heteroscedastic model in \cite{kendall2019geometry} to formulate our reconstruction loss $\mathcal{L}_h^{rec}$ and measure the data-dependent reconstruction uncertainty, where L1 norm is adopted due to its robustness to outlying residuals:
    \begin{equation}
        \mathcal{L}_h^{rec}=\sum \frac{\sqrt 2}{\boldsymbol{\sigma}}||\mathbf{y}_h-\mathbf{\hat{y}}_h||+\log \boldsymbol{\sigma}.
    \end{equation}
    
    Based on the prior information on structured urban terrains, we introduce the unsupervised total variation loss $\mathcal{L}_h^{tv}$ to remove the noises on the surfaces while preserving the edges in the generated height maps:
    \begin{equation}
        \mathcal{L}_h^{tv}=\sum \sum_{j,k} |\hat{y}_{(h)\,j+1,k}-\hat{y}_{(h)\,j,k}|+|\hat{y}_{(h)\,j,k+1}-\hat{y}_{(h)\,j,k}|,
    \end{equation}
    where $j$, $k$ are map indices in rows and columns.
    
    Both the edge and the height map generation additionally adopt the adversarial loss $\mathcal{L}^{adv}$ and the feature matching loss $\mathcal{L}^{fm}$ calculated from the respective discriminators. The generation adversarial loss is defined as:
    \begin{equation}
        \mathcal{L}^{adv}=\sum \log(1-\mathcal{D}(\mathbf{\hat{y}})).
    \end{equation}
    Similar to \cite{nazeri2019edgeconnect}, the feature matching loss is calculated by comparing the feature maps from multiple intermediate convolutional blocks $\mathcal{D}^{(l)}$ in the discriminators:
    \begin{equation}
        \mathcal{L}^{fm}=\sum ||\mathcal{D}^{(l)}(\mathbf{y})-\mathcal{D}^{(l)}(\mathbf{\hat{y}})||.
    \end{equation}
    
    The overall optimization objective of the generation model is the weighted sum of all the above terms, where the generated edge $\mathbf{\hat{y}}_e$, height map $\mathbf{\hat{y}}_h$, and the map of log variance $\log \boldsymbol{\sigma}$ are jointly optimized:
    \begin{equation}
        \mathcal{L}^{total}=\mathbf{w}^T[\mathcal{L}^{bce}_e, \mathcal{L}^{adv}_e, \mathcal{L}^{fm}_e, \mathcal{L}^{rec}_h, \mathcal{L}^{tv}_h, \mathcal{L}^{adv}_h, \mathcal{L}^{fm}_h]^T.
    \end{equation}

\section{Experiments}
    \subsection{Implementation Details} \label{sec:imple}
    \subsubsection{Setup}
    We adopt CoppeliaSim \cite{coppeliaSim} to collect the dataset using the approach described in Section \ref{sec:simu_env}.
    The final dataset contains $\num{40}$k valid samples with an average observation ratio of $\SI{60}{\percent}$.
    The train, validation, and evaluation split ratios are $\num{0.7}$, $\num{0.1}$, and $\num{0.2}$ respectively. 
    Both the simulation and the real-world experiments adopt the same network model trained solely on this simulation dataset.
    
    In simulation experiments, all the mapping approaches are evaluated using $\qtyproduct[product-units=single]{5 x 5}{\metre}$ patches in \textit{urban} environments that contain regions of \textit{stairs}, \textit{slopes} as well as other \textit{obstacles} (boxes and irregular steps).
    We additionally conduct the experiment on $\num{2}$k samples of irregular \textit{hills} with rough surfaces (the second row of Fig. \ref{fig:baseline_vis}), which are quite different from the urban scenarios in our training dataset, to evaluate the generalization capacity of our approach.
    
    For real-world experiments, we specifically focus on the mapping performance on stairs, as it is frequently used for mapping evaluation in \cite{hoeller2022neural,elevationcupy,stolzle2022reconstructing}, and the reconstruction quality of stairs can largely affect the downstream tasks.
    We control a Jueying Mini \cite{Jueying} quadrupedal robot to approach or walk up $5$ stairs with various heights and steepness. The robot receives sparse point clouds from a RoboSense RS-LiDAR-16 at $\SI{10}{\hertz}$ to observe the environment. 
    The robot odometry is obtained using a modified version of \ac{NDT} \cite{magnusson2009three} which continuously compares the current LiDAR frame with a local point cloud map containing $\num{5}$ nearest frames.
    All the mapping approaches are evaluated on $400$ patches ($\qtyproduct[product-units=single]{2 x 2}{\metre}$) on real-world stairs.
    
    To obtain the ground truth maps in the real world, we collect dense point clouds using multiple point frames from an Ouster OS1 LiDAR with $\num{128}$ channels. The poses of these frames are globally optimized using FAST-LIO \cite{xu2021fast}. 
    Next, we adopt VDBFusion \cite{vdbfusion} which performs well with dense input and accurate odometry to reconstruct the terrain structures. 
    We finally rasterize the results as the ground truth height maps. 
    In each experiment, the relative transformation between the ground truth and the generated map origin is obtained by performing \ac{ICP} on the initial sparse observation frame and the dense point cloud. 
    
    \subsubsection{Baseline approaches} \label{sec:baseline}
    We compare the performance of our approach (N.D.E.M.) with the CuPy implementation of elevation mapping \cite{elevationcupy} which provides the raw elevation maps (E.M.C.) and the results after using the Telea \cite{telea2004image} inpainting method (E.M.C. Telea).
    We also adopt VDBFusion to incrementally reconstruct the terrain using incoming sparse observations and rasterize the results for comparison. 
    
    \subsubsection{Ablation study} \label{sec:ablation}
    For ablation study, we train our model:
    \begin{itemize}
        \item without uncertainty estimation (ours w/o unc.), to show the effect of introducing Bayesian learning;
        \item without the edge branch (ours w/o edge.), to show the effect of adding the auxiliary edge generation task;
        \item without using total variation loss (ours w/o tv.), to show the influence of this regularization term;
        \item using valid height and variance from \cite{elevationcupy} after the ``height cell update" as model inputs (ours w/o feat.), to present the importance of additional statistical features.
    \end{itemize}
    The models are trained and evaluated on three random seeds.
    
    \subsubsection{Evaluation metrics} \label{sec:metric}
    The \ac{MAE} is widely adopted to measure the accuracy of pixel values.
    In addition, we calculate the gradient of height w.r.t. the $x$ and $y$ directions and introduce the \ac{MGD} which is the averaged L2 norm of the gradient differences between the ground truths and the generation results to measure the capability of maintaining details at flat surfaces and edges.
    As it is not quite meaningful to consider the accuracy of grids in a heavily occluded region, we mask a grid if the observation ratio of a $\qtyproduct[product-units=single]{1 x 1}{\metre}$ patch around it is less than $\SI{50}{\percent}$ and obtain the masked results (mMAE and mMGD) to measure the reconstruction accuracy.
    We also calculate the \ac{PSNR} and \ac{SSIM} without applying the masks to further evaluate the noise level and visual similarity.
    For the approaches without a guarantee for dense maps (E.M.C. and VDBFusion), we apply the minimum filter in \cite{elevationcupy,jenelten2022tamols} on their results to fill in the empty region for calculation.

    \subsection{Simulation Results}
    Table \ref{tab:simu} presents the mapping performance in simulated \textit{urban} and \textit{hills} scenarios. 
    Our models achieve higher mapping accuracy in height values and their gradients compared with the traditional approaches.
    The high \ac{PSNR} and \ac{SSIM} values indicate the effectiveness of our models in generating high-quality maps and rational terrain structures.
    Even for \textit{hills} that are not contained in our training dataset, our models still present high mapping performance, indicating their strong generalization capabilities.
    By introducing Bayesian learning, our full version (N.D.E.M.) achieves better performance (compared with ours w/o unc.).
    This may be because the uncertainty term can guide the model to focus on recovering the regions with enough information and return a high $\boldsymbol{\sigma}$ to lower the importance of heavily occluded regions in the reconstruction loss.
    In addition, adopting the edge generation branch in training and using our statistical point features can assist to recover the terrain features and further improves the mapping accuracy (compared with ours w/o edge. and ours w/o feat.) in \textit{urban} scenarios.
    While ours w/o edge. performs better in \textit{hills} scenarios compared with other models.
    
    \renewcommand{\arraystretch}{1.2}
    \renewcommand\tabcolsep{2.3pt}
    \begin{table}[t]
    \scriptsize
    \caption{Elevation Mapping Performance on Simulated Urban Terrains and Hills}
    \centering
    \begin{threeparttable}
    \begin{tabular}{l|cccc} 
    \toprule
    \footnotesize \textit{Urban}    & \multicolumn{1}{c}{\footnotesize mMAE $\downarrow$} & \multicolumn{1}{c}{\footnotesize mMGD $\downarrow$} & \multicolumn{1}{c}{\footnotesize PSNR $\uparrow$} & \multicolumn{1}{c}{\footnotesize SSIM $\uparrow$} \\ 
    \hline
    \footnotesize E.M.C.            & $3.56$                 & $0.58$                  & $57.6$                & $0.685$                  \\
    \footnotesize E.M.C. Telea      & $2.62$                 & $0.34$                  & $64.1$                & $0.751$                  \\
    \footnotesize VDBFusion         & $3.17$                 & $0.50$                  & $61.0$                & $0.695$                  \\
    \footnotesize N.D.E.M.(ours)    & $\mathbf{2.10}\pm0.07$ & $\mathbf{0.26}\pm0.004$ & $\mathbf{66.8}\pm0.2$ & $0.738\pm0.012$          \\
    \footnotesize ours w/o unc.     & $2.82\pm0.08$          & $0.28\pm0.005$          & $66.3\pm0.4$          & $0.649\pm0.022$          \\
    \footnotesize ours w/o edge.    & $\mathbf{2.10}\pm0.07$ & $0.27\pm0.006$          & $66.1\pm0.2$          & $0.728\pm0.008$          \\
    \footnotesize ours w/o tv.      & $2.12\pm0.10$          & $0.30\pm0.003$          & $64.4\pm0.2$          & $\mathbf{0.762}\pm0.008$ \\
    \footnotesize ours w/o feat.    & $2.20\pm0.05$          & $0.29\pm0.003$          & $65.3\pm0.1$          & $0.726\pm0.006$          \\
    \toprule
    \footnotesize \textit{Hills}    & \multicolumn{1}{c}{\footnotesize mMAE $\downarrow$} & \multicolumn{1}{c}{\footnotesize mMGD $\downarrow$} & \multicolumn{1}{c}{\footnotesize PSNR $\uparrow$} & \multicolumn{1}{c}{\footnotesize SSIM $\uparrow$} \\ 
    \hline
    \footnotesize E.M.C.            & $6.87$                 & $1.11$                  & $56.0$                & $0.461$                  \\
    \footnotesize E.M.C. Telea      & $3.78$                 & $0.45$                  & $70.0$                & $0.746$                  \\
    \footnotesize VDBFusion         & $4.84$                 & $0.59$                  & $60.8$                & $0.614$                  \\
    \footnotesize N.D.E.M.(ours)    & $2.37\pm0.07$          & $0.41\pm0.002$          & $73.2\pm0.2$          & $0.831\pm0.002$          \\
    \footnotesize ours w/o unc.     & $3.06\pm0.09$          & $0.43\pm0.002$          & $71.4\pm0.2$          & $0.821\pm0.003$          \\
    \footnotesize ours w/o edge.    & $\mathbf{2.21}\pm0.12$ & $\mathbf{0.40}\pm0.001$ & $\mathbf{73.9}\pm0.2$ & $\mathbf{0.836}\pm0.006$ \\
    \footnotesize ours w/o tv.      & $2.48\pm0.06$          & $0.42\pm0.002$          & $72.1\pm0.3$          & $0.823\pm0.001$          \\
    \footnotesize ours w/o feat.    & $2.88\pm0.07$          & $0.41\pm0.001$          & $72.1\pm0.1$          & $0.814\pm0.002$          \\
    \bottomrule
    \end{tabular}
    \footnotesize
    \begin{tablenotes}
        \item See Section \ref{sec:imple} for details of the evaluated approaches and the evaluation metrics. Our models are evaluated over 3 random seeds.
    \end{tablenotes}
    \end{threeparttable}
    \label{tab:simu}
    \vspace{-2.0em}
    \end{table}
    \renewcommand{\arraystretch}{1.0}
    
    Table \ref{tab:simu_terrs} shows the terrain-wise results in simulated \textit{urban} scenarios.
    Our models perform better than the traditional approaches on all kinds of evaluated terrains.
    Our full version (N.D.E.M.) achieves higher mapping performance on \textit{stairs} and \textit{obstacles}, while also having satisfactory results on \textit{slopes}.
    This indicates that the auxiliary edge branch, the total variant regularization, and our representation of point features can assist to recover structured terrains.
    All the approaches present less satisfactory results for \textit{obstacles}, probably because \textit{obstacles} might be irregular and normally have more occlusions, which makes the inpainting difficult.
    
    \renewcommand{\arraystretch}{1.2}
    \renewcommand\tabcolsep{3.8pt}
    \begin{table*}[t]
    \scriptsize
    \caption{Elevation Mapping Performance in Simulation on Different Types of Urban Terrains}
    \centering
    \begin{threeparttable}
    \begin{tabular}{l|ccc|ccc|ccc} 
    \toprule
                              & \multicolumn{3}{c|}{\footnotesize \textit{Stairs}}  & \multicolumn{3}{c|}{\footnotesize \textit{Slopes}}  & \multicolumn{3}{c}{\footnotesize \textit{Obstacles}} \\
    \cline{2-10}
    \footnotesize Methods         & \multicolumn{1}{c}{\footnotesize mMAE $\downarrow$} & \multicolumn{1}{c}{\footnotesize mMGD $\downarrow$} & \multicolumn{1}{c|}{\footnotesize PSNR $\uparrow$}
                                  & \multicolumn{1}{c}{\footnotesize mMAE $\downarrow$} & \multicolumn{1}{c}{\footnotesize mMGD $\downarrow$} & \multicolumn{1}{c|}{\footnotesize PSNR $\uparrow$}
                                  & \multicolumn{1}{c}{\footnotesize mMAE $\downarrow$} & \multicolumn{1}{c}{\footnotesize mMGD $\downarrow$} & \multicolumn{1}{c}{\footnotesize PSNR $\uparrow$} \\
    \hline
    \footnotesize E.M.C.          & $10.65$                & $1.54$                 & $35.2$                & $2.95$                 & $0.50$                  & $46.7$                 
                                  & $14.97$                & $1.71$                 & $54.2$                \\
    \footnotesize E.M.C. Telea    & $5.23$                 & $0.85$                 & $46.8$                & $2.68$                 & $0.40$                  & $53.2$                 
                                  & $8.91$                 & $1.40$                 & $\mathbf{64.7}$       \\
    \footnotesize VDBFusion       & $9.03$                 & $1.37$                 & $39.3$                & $3.04$                 & $0.40$                  & $49.3$                 
                                  & $17.28$                & $1.76$                 & $58.3$                \\
    \footnotesize N.D.E.M.(ours)  & $\mathbf{3.34}\pm0.05$ & $\mathbf{0.68}\pm0.01$ & $\mathbf{47.9}\pm0.1$ & $1.91\pm0.10$          & $\mathbf{0.26}\pm0.005$ & $52.7\pm0.2$           
                                  & $\mathbf{8.83}\pm0.06$ & $\mathbf{1.33}\pm0.02$ & $63.6\pm0.3$          \\
    \footnotesize ours w/o unc.   & $4.80\pm0.08$          & $0.72\pm0.05$          & $47.0\pm0.5$          & $2.36\pm0.08$          & $\mathbf{0.26}\pm0.003$ & $52.0\pm0.3$           
                                  & $10.13\pm0.08$         & $1.35\pm0.05$          & $62.6\pm0.6$          \\
    \footnotesize ours w/o edge.  & $3.47\pm0.06$          & $0.71\pm0.02$          & $\mathbf{47.9}\pm0.1$ & $\mathbf{1.50}\pm0.07$ & $\mathbf{0.26}\pm0.007$ & $\mathbf{53.6}\pm0.2$  
                                  & $9.78\pm0.06$          & $1.36\pm0.03$          & $63.0\pm0.5$          \\
    \footnotesize ours w/o tv.    & $4.49\pm0.12$          & $0.82\pm0.02$          & $47.0\pm0.2$          & $1.89\pm0.08$          & $0.28\pm0.004$          & $53.0\pm0.1$           
                                  & $8.84\pm0.11$          & $1.42\pm0.02$          & $63.2\pm0.3$          \\
    \footnotesize ours w/o feat.  & $5.14\pm0.07$          & $0.81\pm0.01$          & $46.5\pm0.1$          & $1.59\pm0.03$          & $\mathbf{0.26}\pm0.002$ & $53.3\pm0.1$           
                                  & $11.00\pm0.07$         & $1.47\pm0.02$          & $61.6\pm0.2$          \\
    \bottomrule
    \end{tabular}
    \footnotesize
    \begin{tablenotes}
        \item See Section \ref{sec:imple} for details of the evaluated approaches and the evaluation metrics. Our models are evaluated over 3 random seeds.
    \end{tablenotes}
    \end{threeparttable}
    \label{tab:simu_terrs}
    \vspace{-1.0em}
    \end{table*}
    \renewcommand{\arraystretch}{1.0}

    \begin{figure}[h]
    \setlength{\abovecaptionskip}{0pt}
    \centering
    \includegraphics[width=3.0in]{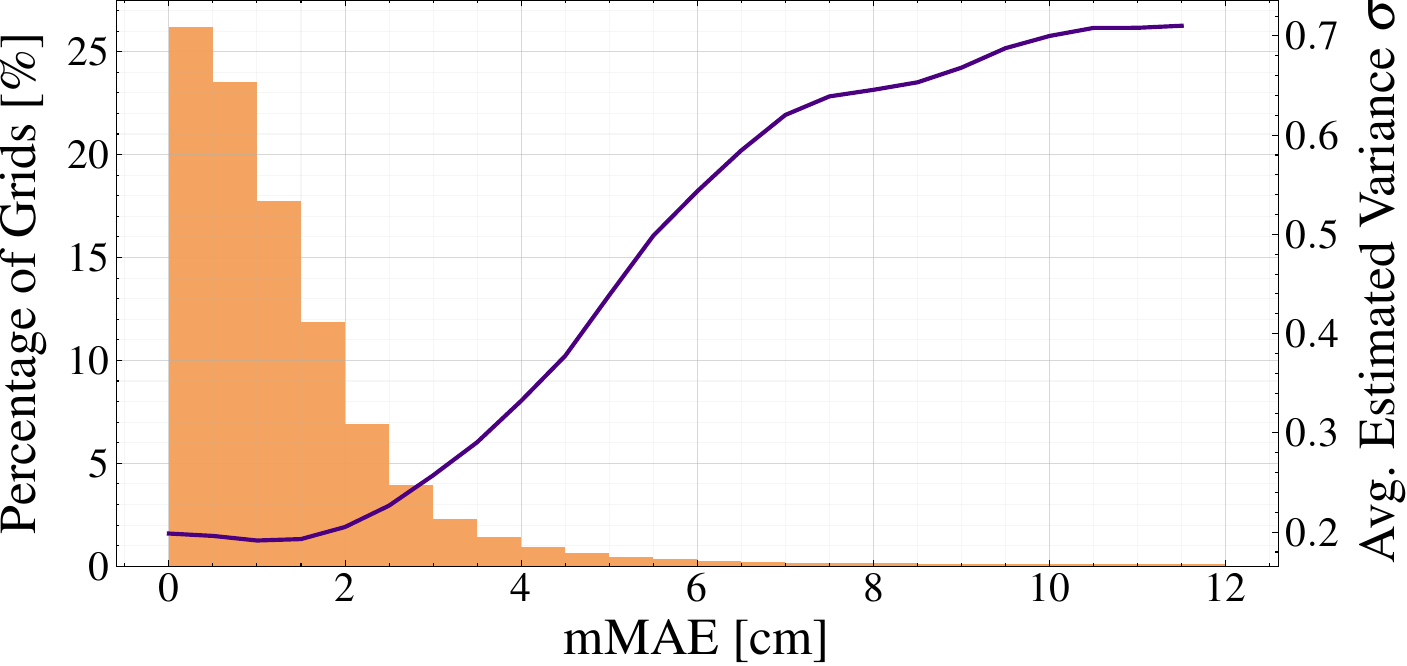}
    \caption{The percentage (orange bar plot) of the grids with different levels of error and the corresponding estimated $\boldsymbol{\sigma}$ values (purple curve). Our approach provides high-quality mapping results that are accurate enough for various downstream tasks. The uncertainty estimations further reflect the confidence of our model on the generated maps.}
    \label{fig:uncertainty_loss_his}
    \vspace{-1.5em} 
    \end{figure}
    
    To intuitively evaluate the accuracy of our approach on map generation and uncertainty estimations, we plot the percentage of different height error levels in simulations and the corresponding averaged $\boldsymbol{\sigma}$ values. 
    The histogram in Fig. \ref{fig:uncertainty_loss_his} shows the percentage of mMAE values among all the \textit{urban} results, which indicates that more than three-quarters of the grids have an error of less than $\SI{2}{\cm}$, and around half of the grids achieve an even higher accuracy (\textless$\SI{1}{\cm}$).
    The curve in Fig. \ref{fig:uncertainty_loss_his} plots the change of $\boldsymbol{\sigma}$ over the unmasked errors considering all the generated height values.
    It shows that $\boldsymbol{\sigma}$ values are proportional to the errors, which successfully reflect the confidence of our model on mapping results.
    
    The first and the second row of Fig. \ref{fig:baseline_vis} visualize two examples in simulated \textit{urban} and \textit{hills} scenarios, where we use $\qtyproduct[product-units=single]{8 x 8}{\metre}$ patches containing multiple urban terrain features for qualitative evaluation.
    We select two of our ablated models (ours w/o unc. and ours w/o feat.) for visualization.
    Our models provide high-quality dense elevation maps in the complex \textit{urban} scenario, and our full version (N.D.E.M.) better recovers the terrain features with straight and sharp edges.
    For \textit{hills} scenarios, our models successfully recover the terrain in a global view, which explains the high \ac{SSIM} scores in table \ref{tab:simu}.
    However, they fail to recover the local features and generate step-like terrains with smooth surfaces.
    
    \subsection{Real-world Performance}
    Table \ref{tab:real} shows the performance on real-world stairs. 
    Both raw E.M.C. with the minimum filter and E.M.C. Telea present poorly probably due to the low observation ratios of the scenes with our sensor setting, and traditional inpainting approaches may require enough observations for satisfactory performance.
    The VDBFusion has an acceptable performance at the expense of memory and computational efficiency as it incrementally constructs a global map, which may introduce a heavy burden to the mobile device.
    Our models generate maps with high accuracy and structural similarity.
    Our full version (N.D.E.M.) achieves the best performance among all the models, presenting the effectiveness of our approach in real-world scenarios.
    
    \renewcommand{\arraystretch}{1.2}
    \renewcommand\tabcolsep{2.5pt}
    \begin{table}[t]
    \scriptsize
    \caption{Elevation Mapping Performance on Real-world Stairs}
    \centering
    \begin{threeparttable}
    \begin{tabular}{l|cccc} 
    \toprule
                     & \multicolumn{1}{c}{\footnotesize mMAE $\downarrow$} & \multicolumn{1}{c}{\footnotesize mMGD $\downarrow$} & \multicolumn{1}{c}{\footnotesize PSNR $\uparrow$} & \multicolumn{1}{c}{\footnotesize SSIM $\uparrow$} \\ 
    \hline
    \footnotesize E.M.C.           & $4.13$                  & $0.69$                   & $73.7$                  & $0.779$                   \\
    \footnotesize E.M.C. Telea     & $3.92$                  & $0.66$                   & $74.1$                  & $0.787$                   \\
    \footnotesize VDBFusion        & $3.41$                  & $0.52$                   & $75.5$                  & $0.839$                   \\
    \footnotesize N.D.E.M.(ours)   & $\mathbf{2.48}\pm0.08$  & $\mathbf{0.39}\pm0.02$   & $\mathbf{77.5}\pm0.4$  & $\mathbf{0.887}\pm0.002$  \\
    \footnotesize ours w/o unc.    & $2.97\pm0.10$           & $0.43\pm0.02$            & $76.6\pm0.5$           & $0.865\pm0.003$           \\
    \footnotesize ours w/o edge.   & $2.80\pm0.09$           & $0.46\pm0.02$            & $76.5\pm0.4$           & $0.859\pm0.002$           \\
    \footnotesize ours w/o tv.     & $2.84\pm0.08$           & $0.41\pm0.03$            & $76.9\pm0.2$           & $0.867\pm0.002$           \\
    \footnotesize ours w/o feat.   & $2.99\pm0.08$           & $0.40\pm0.02$            & $76.6\pm0.2$           & $0.876\pm0.003$           \\
    \bottomrule
    \end{tabular}
    \footnotesize
    \begin{tablenotes}
        \item See Section \ref{sec:imple} for details of the evaluated approaches and the evaluation metrics. Our models are evaluated over 3 random seeds.
    \end{tablenotes}
    \end{threeparttable}
    \label{tab:real}
    \vspace{-1.5em}
    \end{table}
    \renewcommand{\arraystretch}{1.0}
    
    We further visualize an example of the mapping results in a real-world scenario on a stair. 
    As shown in the third row of Fig. \ref{fig:baseline_vis}, E.M.C. generates a noisy map with holes. 
    Although E.M.C. Telea can provide dense mapping results, it fails to recover the map structures from noises and the inpainting is inaccurate if there is not enough observation.
    VDBFusion successfully reconstructs the stair structures for the first several steps.
    However, it still cannot provide dense results and the performance is unsatisfactory for the former steps. 
    Our approach (N.D.E.M.) provides high-quality dense mapping results with clean and sharp edges compared with traditional approaches and the models for ablation study.
    
    \subsection{Run-time Analysis}
    We deploy a Python implementation of the proposed elevation mapping framework on an Nvidia Jetson AGX Xavier. 
    The point pre-processing stage containing all the operations in Section \ref{sec:features} is accelerated using GPU through CuPy element-wise kernels.
    It takes only around $\SI{1.6}{\milli\second}$ to pre-process a point frame from a LiDAR with $\num{16}$ channels ($20$k to $30$k points) and generate the network input.
    It still takes a short time ($\SI{4.7}{\milli\second}$) if we increase the number of points to $1$M.
    As the generation model is fully convolutional, it can deal with input feature maps of diverse sizes and generate maps of the corresponding shape.
    The model inference time is irrelevant to the size of point clouds after the pre-processing stage but related to the size of the mapping region.
    The model takes around $\SI{24}{\milli\second}$ to generate $\qtyproduct[product-units=single]{3.2 x 3.2}{\metre} \ (\qtyproduct[product-units=single]{80 x 80}{\px})$ local patches for locomotion as done in \cite{hoeller2022neural} and uses $\SI{75}{\milli\second}$ to generate $\qtyproduct[product-units=single]{12 x 12}{\metre} \ (\qtyproduct[product-units=single]{300 x 300}{\px})$ maps for cost-optimal navigation on complex terrains as required in \cite{yang2021real}.
    The whole pipeline performs mapping at $\SI{10}{\hertz}$, achieving real-time performance on the mobile computation platform.
    
    \subsection{Benefits to Downstream Tasks}
\begin{figure}[h]
    \setlength{\abovecaptionskip}{0pt}
    \centering
    \includegraphics[width=3.2in]{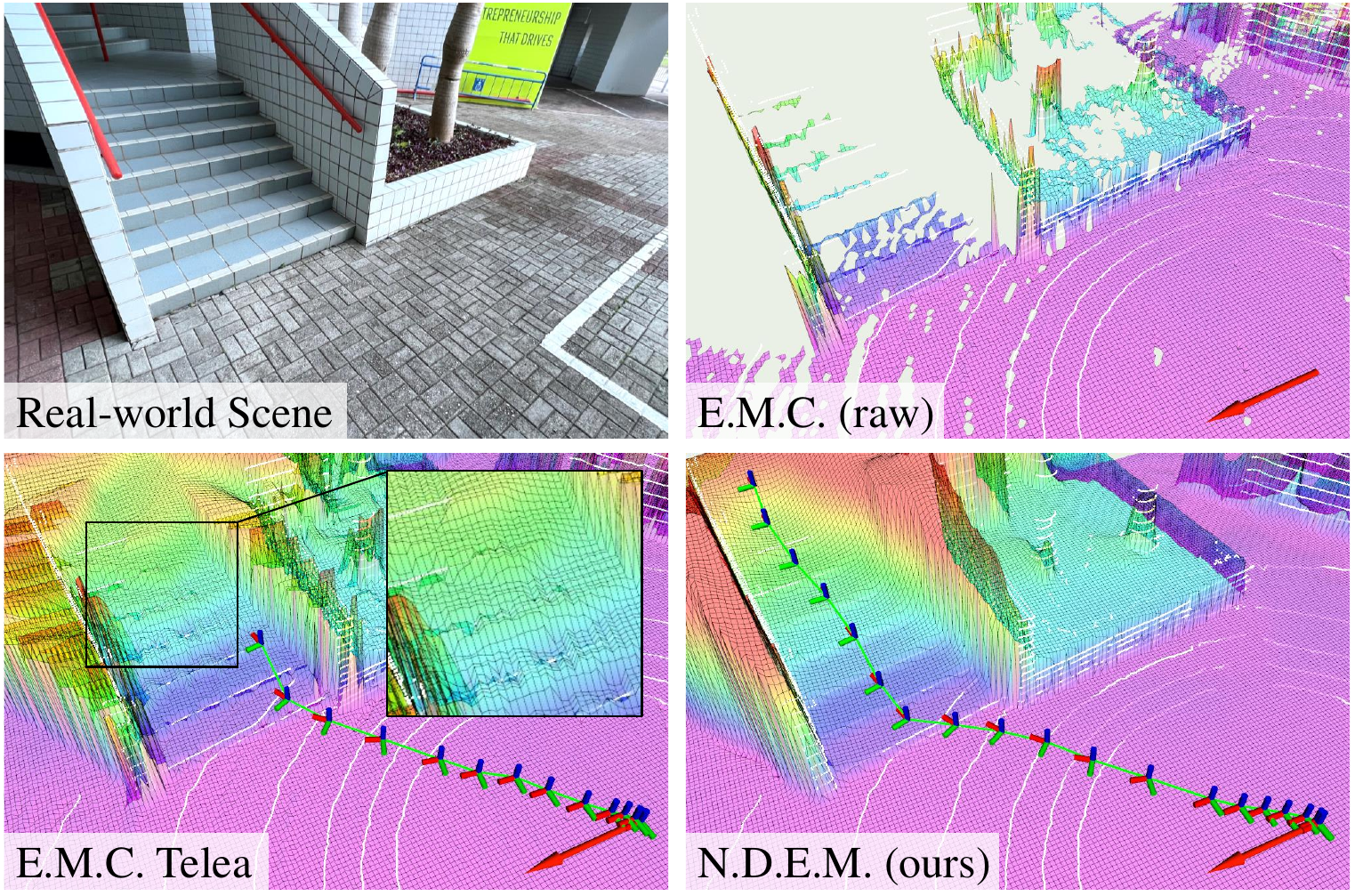}
    \caption{Path planning experiment with the online mapping results on real-world stairs. E.M.C. Telea generates rough steps and wrong estimations of the step height, resulting in the failure of planning. Our approach successfully recovers the stair structure at a glance, presenting strong capabilities in urban terrain reconstruction with extremely sparse observation.}
    \label{fig:plan}
    \vspace{-1.0em}
    \end{figure}

    We then conduct path planning on real-world stairs using the approach in \cite{yang2021real} which requires dense elevation maps for optimal path planning on complex terrains.
    As shown in Fig. \ref{fig:plan}, both E.M.C. Telea and our approach provide online dense elevation maps when the robot approaches the stair. 
    However, E.M.C. Telea generates rough step surfaces and wrongly estimates the height of stairs under extremely sparse observations, resulting in the failure of the planning task.
    Our approach successfully reconstructs the stair once the LiDAR observes the steps.
    The strong capability of our approach in reconstructing urban terrain features at a glance assists the navigation tasks to achieve better performance.
    
\begin{figure}[h]
    \setlength{\abovecaptionskip}{0pt}
    \centering
    \includegraphics[width=3.3in]{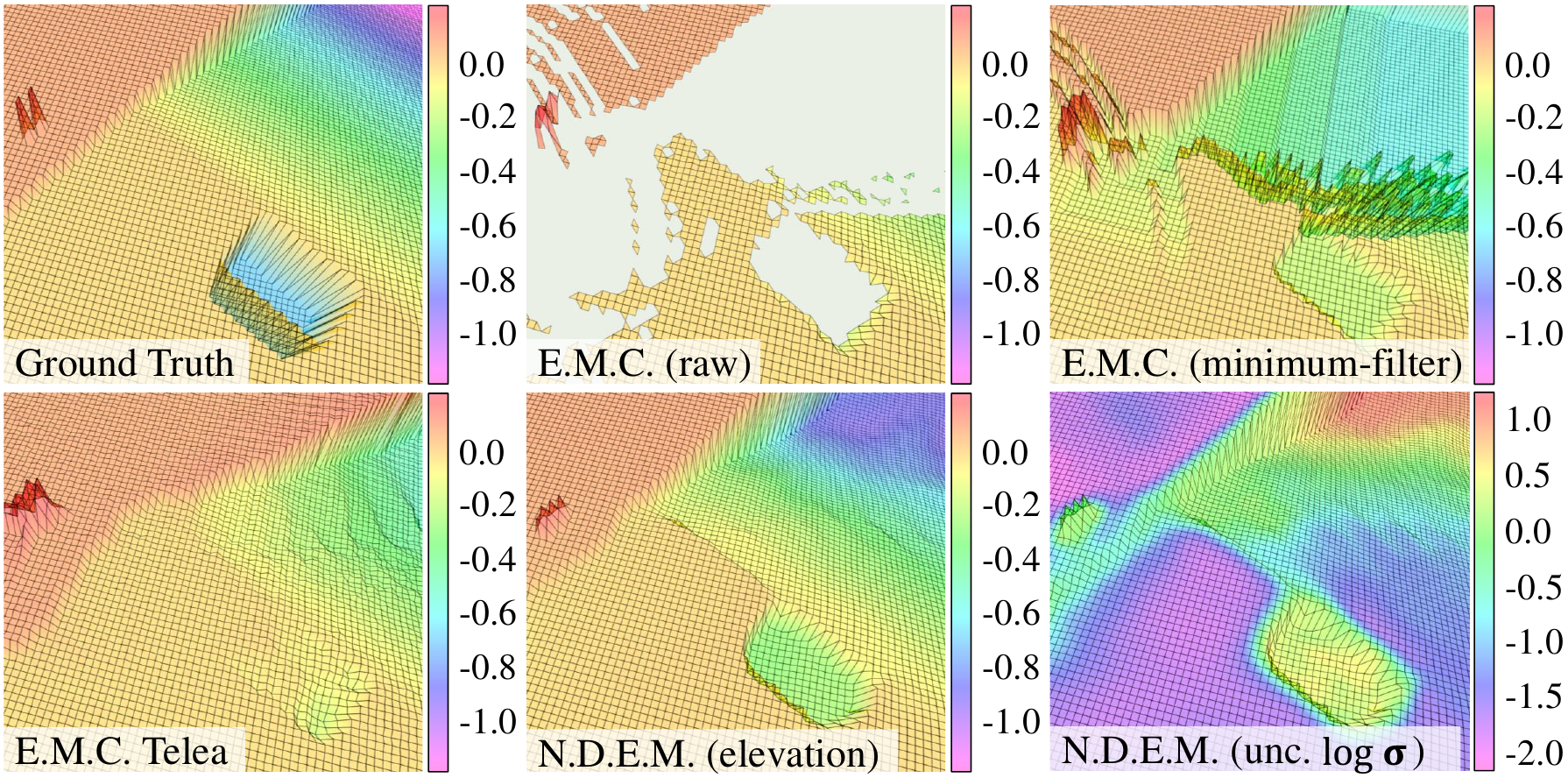}
    \caption{When encountering the occluded regions, rather than just filling in the region as done by the traditional inpainting approaches, our mapping module returns high uncertainty at the occluded region, informing the downstream tasks of the potential danger.}
    \label{fig:hole}
    \vspace{-1.0em}
    \end{figure}

    Adopting the uncertainty estimations in the downstream tasks may help to ensure safe and robust behaviors of the robot. 
    Fig. \ref{fig:cover}(c) visualizes the uncertainty values on the real-world stairs. 
    Our model returns a higher uncertainty at the edges of the stairs than the flat surfaces, which may assist the locomotion algorithms to step feet onto safe regions and prevent slips resulting from the terrain reconstruction inaccuracy.
    Also, when there is a hole in the ground (Fig. \ref{fig:hole}), traditional inpainting approaches (E.M.C. with minimum filter and E.M.C. Telea) \cite{elevationcupy} just fill in that region to make it a surface, resulting in potential danger to navigation tasks.
    Although our approach cannot provide accurate height values inside the hole due to occlusion, it returns high reconstruction uncertainty that can inform the navigation module for safe behaviors.
    In addition, our approach can generate the slope based on the information of adjacent grids and returns higher uncertainty values as the occlusion gets stronger.
    
\section{Conclusion}
    We proposed a learning-based dense elevation mapping framework for online urban terrain reconstruction. 
    It maintains high reconstruction quality and accuracy in both simulation and real-world scenarios.
    The proposed data pre-processing method and the map generation model show robust performance that can recover detailed urban terrain structures from sparse and noisy LiDAR observations, while also achieving high efficiency to enable real-time elevation mapping on mobile devices.
    The provided uncertainty estimations also have the potential to assist the downstream tasks for safer and more robust performance.
    
    There are also several limitations of our approach.
    Our model overfits the urban scenarios and fails to recover the detailed features of irregular terrains or obstacles.
    The representation approach of point features needs to be improved to deal with overhanging obstacles and dynamic objects.
    In addition, the pipeline has no specific design to compensate for the drifts and it assumes the drifts are negligible in local mapping scenarios.
    Currently, a height drift of several centimeters will affect the stability of the point features and lead to degraded mapping performance.
    Jointly optimizing the map and the odometry might be a future direction to solve this problem.

                                
\section*{Acknowledgment}
We would like to thank Jianhao Jiao, Peng Yun and Xiangcheng Hu for their valuable suggestions.

\bibliography{reference} 
\bibliographystyle{ieeetr}

\end{document}